\documentclass[11pt]{article}

\usepackage[final]{coling}

\usepackage{times}
\usepackage{latexsym}
\usepackage{subcaption}
\usepackage{amsmath,amsfonts,amssymb}
\usepackage{breqn}
\usepackage{tabularx}
\usepackage{multirow}
\usepackage{graphicx}
\usepackage{color}
\usepackage{tcolorbox}
\usepackage{CJK}
\usepackage{adjustbox}
\usepackage{xcolor}
\usepackage{colortbl}
\usepackage{multicol}
\usepackage{vwcol} 
\newsavebox{\algleft}
\newsavebox{\algright}
\usepackage{bm}
\usepackage{booktabs}
\usepackage{ctable} 
\usepackage{amsmath,stackengine}
\usepackage[linesnumbered,ruled,vlined]{algorithm2e}
\usepackage{times}
\usepackage{latexsym}
\usepackage{bbm}
\usepackage[utf8]{inputenc} 
\usepackage[T1]{fontenc}    
\usepackage{hyperref}       
\usepackage{url}            
\usepackage{booktabs}       
\usepackage{amsfonts}       
\usepackage{nicefrac}       
\usepackage{microtype}      
\usepackage{xcolor}         
\usepackage{bm}
\usepackage{microtype}

\usepackage{latexsym}

\usepackage[T1]{fontenc}

\usepackage[utf8]{inputenc}

\usepackage{microtype}

\usepackage{inconsolata}

\usepackage{dblfloatfix}

\title{Continual Learning Using Only Large Language Model Prompting \thanks{To Appear in COLING-2025}}

\author{Jiabao Qiu$^1$, Zixuan Ke$^2$, Bing Liu$^1$ \\
  $^1$Department of Computer Science, University of Illinois Chicago \\
  $^2$Salesforce AI Research\\
  $^1$\texttt{\{jqiu23, liub\}@uic.edu} \\
$^2$\texttt{zixuan.ke@salesforce.com}\\}

\begin{document}
\maketitle

\begin{abstract}


We introduce \textit{CLOB}, a novel \textit{continual learning} (CL) \textit{paradigm} wherein a large language model (LLM) is regarded as a black box. Learning is done incrementally via only verbal prompting. 
CLOB does not fine-tune any part of the LLM or add any trainable parameters to it. 
It is particularly suitable for LLMs that are accessible via APIs. 
We also propose a new CL technique, called CIS, based on \textit{incremental summarization} that also overcomes the LLM's input length limit. Experiments show CIS outperforms baselines by a very large margin. 


\end{abstract}

\section{Introduction}
\label{sec.intro}



{\color{black} Continual learning (CL) learns
a sequence of tasks incrementally~\cite{chen2018lifelong}. Once a task is learned, its training data is discarded.  
{\color{black}Existing CL work in NLP mainly fine-tunes language model (LM) parameters or uses adapters or variants to adapt the LM~\cite{ke2022continual}.} In learning a new task, the system must update the network parameters without seeing the old task data. This may corrupt the knowledge learned from old tasks, causing \textit{catastrophic forgetting} (CF). Further, since a large number of training samples, expensive training, and access to the LM parameters are needed, it is limited to ``small'' LMs. 

Recently, large language models (LLMs) such as GPT4 \cite{openai2023gpt4} and Gemini \cite{Google2024} have revolutionized NLP, but they are typically accessed only through APIs. In these ``black-boxes''\footnote{We use the term \textit{black-box} because, as users, we can only interact with LLMs using verbal prompts or via an API.} LLMs, {\color{black}the users typically use prompts that include few-shot in-context examples and instructions to ask the LLMs to perform their tasks. 


CL with black-box LLMs is clearly important. However, to our knowledge, CL has not been explored in this context.} Our \textbf{first contribution} is thus to propose a new CL paradigm for this context, called \textbf{CLOB} (\textbf{C}ontinual \textbf{L}earning \textbf{O}ver \textbf{B}lack-box LLMs). The user works with the LLM using only verb prompts with few-shot in-context examples and instructions. The traditional \textbf{\textit{parameter-based CF}} caused by parameter updating \textit{disappears}, but a new \textbf{\textit{prompt-based CF}} \textit{appears}. 

This paper works in the \textit{class-incremental learning} (CIL) setting of CL. In CIL, each task consists of one or more classes to be learned. In testing, no task identification information is given.
To make CIL more realistic, we allow the arrivals of the training data from different tasks to intertwine, called \textbf{\textit{blurry task boundaries}}. That is, when a new task arrives, only a portion of its training data is available and the rest of the labeled samples of the task may come at any time later \cite{koh2021online}.
{\color{black}This is referred to as \textbf{\textit{online}} or \textbf{\textit{streaming} CIL}, where the training data arrives as a continuous stream, and each training example is seen only once by the system \cite{guo2022online}.} 

Since an LLM has a \textbf{maximum input length} or \textbf{token} \textbf{limit}, it restricts the number of in-context examples that can be used in a prompt. This poses a major challenge for CIL because it needs to learn more and more tasks/classes. 
Thus, the ability to learn and store a minimum amount of knowledge for each class and to incrementally \textbf{\textit{update the knowledge on the fly}} when additional data of old tasks becomes available is critical. 
This update also needs to ensure that the knowledge learned previously is not forgotten. Unlike existing approaches, in CLOB, we must do all these by using verbal prompts without touching any network parameters. 

Our \textbf{second contribution} is to propose a simple and novel CIL method for CLOB, called {\color{black}CIS (\textit{in-context} \textit{\textbf{C}}L via }\textit{\textbf{I}ncremental \textit{\textbf{S}}ummarization)}. CIS leverages the summarization capabilities of LLMs to encapsulate the knowledge about each class in a summary and incrementally learn and update the summary when some data from some old tasks come. CIS can also deal with the token limit issue for LLMs. Experimental results show significant accuracy gains compared to baselines. 

}

\noindent
\textbf{Related Work.} Overcoming CF is a key goal of CL. 
There are many existing approaches, e.g., 
\textit{regularization-based approaches}~\cite{Kirkpatrick2017overcoming,li-etal-2022-overcoming,DBLP:conf/naacl/LiuUS19}, 
\textit{replay-based approaches}~\cite{Rebuffi2017,
DBLP:journals/corr/abs-2108-04445,scialom2022continual,qin2022elle,huang2021continual} and  \textit{parameter isolation based approaches} \cite{ke2021achieving,ke2023continual,Serra2018overcoming,gururangan2021demix,qin2022elle,zhu2022continual,DBLP:conf/acl/GengYXSX020,madotto2020continual}. 
{\color{black}Recent research in NLP also used parameter-tuning \cite{zhao2024saptsharedattentionframework} or a rehearsal-free modular \cite{wang2024rehearsalfreemodularcompositionalcontinual} to help knowledge transfer among tasks.} 
 A data-efficient CL paradigm for fine-tuning LLMs \cite{guo2024qtuningqueuebasedprompttuning} and a prompt tuning-based CL method \cite{wang2024inscldataefficientcontinuallearning} are also proposed.
Reflexion \cite{shinn2023reflexion} and LLM as optimizer \cite{yang2023large} update previous knowledge without knowledge retention. Voyager \cite{wang2023voyager} adds new skills to a library and retrieves it for each new task. We are different as we treat LLMs as black boxes and use only verb prompting for CL. 


Many few-shot and instruction prompting techniques are proposed in \cite{wei2022chain,zhou2022least, khot2022decomposed,yao2023tree,hao2023reasoning}.
However, they do not do prompting for CL. 

\begin{figure*}[t]
\center
\includegraphics[width=0.8\textwidth]{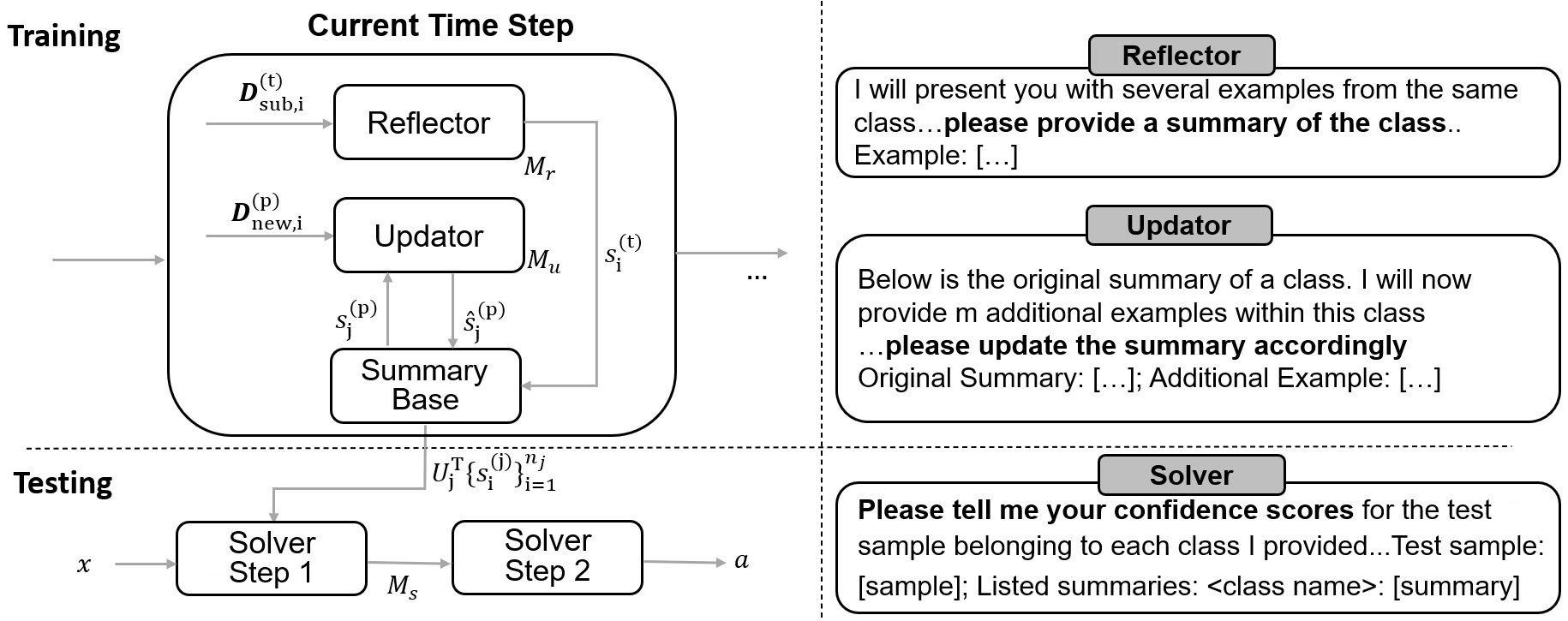}
\caption{\textbf{(1) Left:} Overview of CIS in CLOB. \textbf{(2) Right:} Prompts used in each component of learning. Full prompts can be found in Appendix~\ref{sec:appendix.prompts}. {\color{black}Some example summaries are given in Appendix~\ref{sec:appendix.examples}. }
}
\centering
\label{fig.overview}
\end{figure*}

\section{Proposed CIS Method for CLOB}
\label{sec.approach}




CIS learns a sequence of tasks 1, ..., $T$ in CIL. Each task $t$ has an input space $\mathcal{X}^{(t)}$, a class label space $\mathcal{Y}^{(t)}$ ($\mathcal{Y}^{(t)} \cap \mathcal{Y}^{(i)} = \emptyset$ for all $i \neq t$), and a training set $\mathcal{D}^{(t)} = \{(x_j^{(t)}, y_j^{(t)})\}_{j=1}^{n^{(t)}}$ drawn \textit{i.i.d.} from
$\mathcal{P}_{\mathcal{X}^{(t)}\mathcal{Y}^{(t)}}$. When task $t$ arrives, only a subset $\mathcal{D}_{\textit{sub}}^{(t)}$ of $\mathcal{D}^{(t)}$ is available for learning. The rest of the training samples $\mathcal{D}_{\textit{rest}}^{(t)}= \mathcal{D}^{(t)} \backslash \mathcal{D}_{\textit{sub}}^{(t)}$ may arrive at any time later. That is why we say that the arrivals of the data from different tasks intertwine, or the task boundaries are \textit{\textbf{blurry}}. 
Our goal is to learn a function $f: \cup^T_{t=1}\mathcal{X}^{(t)} \rightarrow \cup^T_{t=1}\mathcal{Y}^{(t)}$ to predict the class label of each test sample $x$. No task identifier is given for a test sample in testing. 

{\color{black}CIS works in online CIL~\cite{guo2022online,DBLP:conf/nips/dAutumeRKY19,wang2020efficient} with \textit{\textbf{no training sample saved}} after being seen. Learning is done incrementally via verbal prompts over a black-box LLM whenever some training samples arrive.}


\vspace{-1mm}
\subsection{CIS System}
Due to the blurry task boundary and {\color{black} stream data}, 
the following three scenarios may occur \textbf{during training} and the system needs to learn from any arrival data immediately and incrementally: 
\textbf{(1)} a new task $t$ arrives with only a subset of the training data $\mathcal{D}_{\textit{sub}}^{(t)}$. The system \textit{\textbf{generates a summary}} of the data in each class and saves the summary. 
\textbf{(2)} Only a set $S$ of new samples from the remaining samples of the old tasks arrives. CIS \textbf{\textit{updates the summary}} of each class in $S$ using samples of the class in $S$. \textbf{(3)} Both (1) and (2) occur. CIS performs the corresponding actions of (1) and (2). Below, we present the summary generator and the summary updator. An overview of CIS is given in Figure~\ref{fig.overview}. 
All prompts used are given in Appendix~\ref{sec:appendix.prompts}.

\vspace{+1mm}
\noindent
\textbf{Summary generator ($M_r$)}: We also call this system the \textit{reflector}. It is used only when a new task arrives.  
Since the data is discarded after it is learned, 
we propose to use the \textit{verbal summary} to represent each class.
Specifically, given the data from a task $t$, $\mathcal{D}_{\textit{sub}}^{(t)}$, the LLM is prompted to generate a summary $s_i^{(t)}$ for each new class $i$ in the task using the data of the class $\mathcal{D}_{\textit{sub}, i}^{(t)}$, which is expected to represent the knowledge of the class,
\begin{equation}
    s_i^{(t)} = M_r(\mathcal{D}_{\textit{sub}, i}^{(t)}).
\end{equation}
The resulting summary $s_i^{(t)}$ is saved. We call all saved summaries the {\color{black}\textbf{\textit{summary base}}}. 

\vspace{+1mm}
\noindent
\noindent\textbf{Summary Updator ($M_u$).}
As \textit{new} data $\mathcal{D}_{\textit{new}}^{(\textit{pre})}$ 
from \textit{pre}vious tasks may come later at any time. The summaries of the 
classes involved in the new data need to be updated. This is done by the Updator. This update can potentially cause forgetting of existing knowledge in the summaries. {\color{black}We call this \textit{prompt-based forgetting}.} Specifically, for each class $j$ (from a previous task $p$) 
involved in the new data $\mathcal{D}_{\textit{new}}^{(\textit{pre})}$, its summary $s_j^{(p)}$ is updated with the new data of class $j$, $\mathcal{D}_{\textit{new, j}}^{(\textit{pre})}$. This incremental summarization is also done via prompts.
\begin{equation}
\hat{s}_j^{(p)} = M_u(s_j^{(p)},\mathcal{D}_{\textit{new, j}}^{(\textit{p})}).
\end{equation}
where $p \in \textit{pre}$. The updated summary, $\hat{s}_j^{(p)}$, replaces the original summary $s_j^{(p)}$ in the {\color{black}summary base}. Some summary examples are given in Appendix~\ref{sec:appendix.examples}. 

\vspace{+2mm}
\noindent
\noindent\textbf{Solver ($M_s$) for Testing.} Due to the prompt length/token limit of LLMs and the fact that a test sample $x$ can be from any learned class,  testing is done in two steps. In Step 1, we divide all classes into multiple chunks.~Each chunk consists of summaries of $k$ classes in the {\color{black}summary base} (no saving of any of their data) that can fit within the length limit of the LLM. We prompt the LLM to generate its confidence that $x$ belongs to each class in the chunk $j$, 
$C_j=\cup_{i=1}^{k}\{s_i\}$,
where $s_i$ is the summary of class $i$ in the chunk (task information is not needed), 
\begin{equation}
\textit{conf}_j = M_s(x,C_j).
\end{equation}
In Step 2, the system first identifies the top $k$ classes with the highest confidence values from all the chunks in Step 1, and then prompts the LLM again with only the resulting $k$ classes, as before, to select the class with the highest confidence for $x$.

\vspace{2mm}
\noindent
{\color{black}\textbf{Theoretical Justification.} It has been shown theoretically that good within-task prediction (WP) and good out-of-distribution (OOD) detection of each task within the tasks that have been learned so far is \textit{necessary} and \textit{sufficient} conditions for good CIL~\cite{kim2022theoretical,kim2023learnability}. In our approach, since we represent each class with its summary in the classification, effectively, each task has only one class at the inference time. Then, WP probability is 1 and CIL depends only on OOD detection of each class. Since each class's summary describes each class's key content, anything that does not belong to the class is dissimilar to the summary, which means the summary helps achieve a good OOD detention effect for each class. Although we don't use OOD detection in the final classification, the LLM is probably making some kind of similarity comparison between each test sample and each class summary.  
}



\begin{table*}[!b]
\renewcommand\thetable{2}
\centering
\scalebox{0.61}{
\begin{tabular}{|l|l|l|l|l|l|l|l|l|l|l|l|l|}
\hline
 & \multicolumn{2}{c|}{\textbf{CIS (Llama)}} & \multicolumn{4}{c|}{\textbf{Joint (Llama)}} & \multicolumn{6}{c|}{\textbf{CL Baselines}} \\ \hline
\multirow{1}{*}{} & \multicolumn{1}{c|}{3/4-Blurry} & \multicolumn{1}{c|}{4/3-Blurry} & \multicolumn{1}{c|}{Zero-shot} & \multicolumn{1}{c|}{Prompting} & \multicolumn{2}{c|}{Fine-tuning} & \multicolumn{2}{c|}{EWC} & \multicolumn{2}{c|}{LAMOL} & \multicolumn{2}{c|}{VAG}  \\ \cline{2-13} 
 & \multicolumn{1}{c|}{7 samples} & \multicolumn{1}{c|}{7 samples} & \multicolumn{1}{c|}{no sample} & \multicolumn{1}{c|}{7 samples} & \multicolumn{1}{c|}{7 samples} & \multicolumn{1}{c|}{full data} & \multicolumn{1}{c|}{7 samples} &  \multicolumn{1}{c|}{full data} & \multicolumn{1}{c|}{7 samples} & \multicolumn{1}{c|}{full data} & \multicolumn{1}{c|}{7 samples} & \multicolumn{1}{c|}{full data} \\ \hline
Banking-77 & 78.78 \tiny{$\pm$1.68} & 79.23 \tiny{$\pm$2.50} & 50.22 \tiny{$\pm$0.00} &  87.92 \tiny{$\pm$0.60} & 69.39 \tiny{$\pm$0.17} & 91.19 \tiny{$\pm$0.08} &  2.14 \tiny{$\pm$0.35} & 9.09 \tiny{$\pm$0.84} &  3.50 \tiny{$\pm$0.04} & 33.43 \tiny{$\pm$0.18} & 36.25 \tiny{$\pm$3.80} & 55.19 \tiny{$\pm$1.54} \\ \hline
CLINC-80 & 91.51 \tiny{$\pm$4.35} & 90.40 \tiny{$\pm$5.46} & 80.67 \tiny{$\pm$0.00} &  95.10 \tiny{$\pm$2.51} & 91.18 \tiny{$\pm$0.46} & 97.92 \tiny{$\pm$0.06} &  1.14 \tiny{$\pm$0.33} & 8.26 \tiny{$\pm$0.76} &  17.60 \tiny{$\pm$0.19} & 52.20 \tiny{$\pm$0.09} & 64.75 \tiny{$\pm$0.69} & 80.68 \tiny{$\pm$0.72} \\ \hline
DBpedia-14 & 92.07 \tiny{$\pm$1.07} & 92.26 \tiny{$\pm$0.76} & 93.36 \tiny{$\pm$0.00} &  90.50 \tiny{$\pm$0.40} & 93.74 \tiny{$\pm$0.11} & 99.00 \tiny{$\pm$0.00} &  6.55 \tiny{$\pm$0.73} & 23.14 \tiny{$\pm$1.55} &  0.70 \tiny{$\pm$0.14} & 28.61 \tiny{$\pm$0.02} & 55.36 \tiny{$\pm$3.30} & 56.58 \tiny{$\pm$1.22} \\ \hline
Reuters-14 & 83.97 \tiny{$\pm$1.08} & 84.61 \tiny{$\pm$1.24} & 92.55 \tiny{$\pm$0.48} & 77.82 \tiny{$\pm$2.99} & 82.64 \tiny{$\pm$0.33} & 92.55 \tiny{$\pm$0.48} & 7.70 \tiny{$\pm$0.70} & 12.79 \tiny{$\pm$0.14} & 0.95 \tiny{$\pm$0.07} & 29.93 \tiny{$\pm$0.17} & 44.08 \tiny{$\pm$0.27} & 58.71 \tiny{$\pm$1.92} \\ \hline
\end{tabular}
}
\caption{\textit{Last Accuracy} of CIS and baselines (2 classes per task for each dataset). CIS's results are copied from Table~\ref{tab:1}.  
}
\vspace{-3mm}
\label{tab:2}
\end{table*}

\section{Experiments}
\label{sec.exp}
\noindent
\textbf{Datasets.} We use four datasets.~\textbf{(1) Banking-77} with 77 classes \cite{casanueva-etal-2020-efficient}, \textbf{(2) CLINC-80} \cite{larson-etal-2019-evaluation} with 80 classes from 10 domains, \textbf{(3) DPpedia-14}~\cite{dbpedia} with 14 classes, and \textbf{(4) Reuters-14}~\cite{reuters} with the top-14 most frequent classes. 
Due to budget constraints and the use of OpenAI's paid GPT-3.5 API, we limited our study to 80 of the 150 classes in the CLINC dataset. However, CIS is capable of handling all 150 classes. Additionally, we focus on classification datasets since CIL is not well-suited for various NLP tasks like summarization, question-answering, and sentiment analysis.


For training, we select 7 random samples per class for all datasets. 7 was chosen to balance the need for variability in our settings with cost considerations. \textit{\textbf{Each task consists of 2 classes}}, except for the last task of Banking-77, which has only one class. For testing, we use 50 samples per class to save money.  
 Tasks are ordered randomly, and the arrival times of new samples are also randomized in the Blurry setting. We perform 3 random runs for each experiment and report the average accuracy.




\textbf{Blurry Setting.}  It is denoted by `$M$ / $N$ \textit{Blurry}', where $M+N = 7$, $M$ is the number of training samples per class used when a task first appeared and $N$ is the number of additional training samples per class that randomly appear later.
{\color{black}Note that although the total number of training samples per class is 7 in this paper due to the budgetary constraint, this blurry setting allows CIS to handle any number of training samples per class and any possible $M$ and $N$ values.}


\textbf{Baselines.}
We compare CIS with several baselines: {\color{black}\textbf{(1)} EWC, a regularization-based algorithm to address CF \cite{doi:10.1073/pnas.1611835114}; \textbf{(2)} LAMOL, a pseudo-replay method that generates data from previous tasks using GPT-2 \cite{sun2020lamol}; \textbf{(3)} VAG, a state-of-the-art parameter-updating CIL method \cite{shao2023classincremental} that fine-tunes the BART \cite{lewis2019bart}.
{\color{black}None of these baselines can work with the blurry setting. Our primary baseline is VAG, as it outperforms other CIL methods. We can only use its original implementation as it does not work with more advanced LLMs like Llama. 

We also include \textbf{(4)} `Joint: zero-shot,' where we use only the class names but no examples in the prompt; \textbf{(5)} `Joint: prompting,' where Joint means to combine the data of all classes in each dataset into a single task and use them in a single prompt; \textbf{(6)} `Joint: fine-tuning,' similar to (5), but we use all {\color{black} available data from all classes in each dataset} to fine-tune the LLM. Joint systems are not CL systems but the upper bounds of CL. 

\textbf{Large language models} (LLMs):
We experimented with three LLMs as black-box models: (1) \textit{Mistral-7B-Instruct-v0.3}, (2) \textit{Llama-3.1-8B-Instruct}, and (3) \textit{GPT-3.5-turbo-instruct}. To ensure comparability, we use the same settings for each LLM. The temperature is set to 0 to make the output deterministic. Each \textbf{summary} is limited to 3 sentences with less than 100 tokens. 

\textbf{Ablations.} \textbf{(1)} Different $M/N$ combinations within CIS;
\textbf{(2)} `$7$ Non-blurry,' where all $7$ samples per class for a task are seen simultaneously, representing the batch CIL approach with clear task boundaries;
{\color{black}\textbf{(3)} {CIS (classify)}, where, in Step 2 of CIS, we ask the LLM to directly output the predicted class rather than the confidence. 



\textbf{Implementation details.} Due to the token limit of GPT-3.5, we split Banking-77 into 3 chunks of sizes 26/26/25 and CLINC-80 into 4 chunks of sizes 20/20/20/20. For the other two datasets, a single chunk suffices. In testing in Step 2, we select \textit{top k} classes for each test sample, with $k=5$. We also tried other $k$ values with $k=5$ being the best. 

{\color{black}
Different prompts may give different results, similar to how different hyper-parameters may affect deep learning models. We choose the most effective from several prompting strategies on dataset CLINC and DBpedia and apply them to all datasets. This shows the robustness of our approach, as we did not finetune the prompts for each dataset. The prompts used are given in Appendix \ref{sec:appendix.prompts}.
}
}

\subsection{Evaluation Results and Analysis}
\vspace{-2mm}

\begin{table}[!h]
\renewcommand\thetable{1}
\begin{center}
\scalebox{0.60}{
\begin{tabular}{|l|l|l|l|l|}
\hline
\multirow{1}{*}{\textbf{CIS Ablations}} & \multicolumn{1}{c|}{3/4-Blurry} & \multicolumn{1}{c|}{4/3-Blurry} & \multicolumn{1}{c|}{5/2-Blurry} & \multicolumn{1}{c|}{7 Non-blurry} \\ \cline{1-5} 
\multicolumn{5}{l}{\textbf{Banking-77}} \\ \hline
CIS (Mistral) & 66.47 \tiny{$\pm$5.88} & 67.42 \tiny{$\pm$3.74}  & 69.54 \tiny{$\pm$4.68}  & 67.91 \tiny{$\pm$1.91} \\ \cline{1-5}
CIS (Llama) & 78.75 \tiny{$\pm$1.68}  & 79.23 \tiny{$\pm$2.50}  & 77.77 \tiny{$\pm$1.49}  & 79.93 \tiny{$\pm$1.74} \\ \cline{1-5}
CIS (GPT) & 85.20 \tiny{$\pm$2.02}  & 85.26 \tiny{$\pm$2.12}  & 85.58 \tiny{$\pm$1.91}  & 85.85 \tiny{$\pm$2.06} \\ \cline{1-5}
CIS (classify - GPT) & \textbf{85.58} \tiny{$\pm$0.57}  & \textbf{85.48} \tiny{$\pm$0.60} &  \textbf{86.05} \tiny{$\pm$0.82}  & \textbf{86.37} \tiny{$\pm$1.45} \\ \hline
\multicolumn{5}{l}{\textbf{CLINC-80}} \\ \hline
CIS (Mistral) & 75.33 \tiny{$\pm$6.75} & 78.71 \tiny{$\pm$6.28}  & 77.54 \tiny{$\pm$7.54}  & 77.17 \tiny{$\pm$8.14} \\ \cline{1-5}
CIS (Llama) & 91.51 \tiny{$\pm$4.35}  & 90.40 \tiny{$\pm$5.46}  & 91.29 \tiny{$\pm$3.61}  & 92.14 \tiny{$\pm$4.51} \\ \cline{1-5}
CIS (GPT) & \textbf{93.88} \tiny{$\pm$3.76}  & \textbf{93.53} \tiny{$\pm$3.84}  & \textbf{93.94} \tiny{$\pm$3.82}  & \textbf{94.22} \tiny{$\pm$4.25} \\ \cline{1-5}
CIS (classify - GPT) & 89.99 \tiny{$\pm$5.37}  & 89.35 \tiny{$\pm$6.14} &  89.43 \tiny{$\pm$4.85}  & 90.93 \tiny{$\pm$5.92} \\ \hline
\multicolumn{5}{l}{\textbf{DBpedia-14}} \\ \hline
CIS (Mistral) & 80.43 \tiny{$\pm$2.83} & 85.32 \tiny{$\pm$0.15}  & 79.29 \tiny{$\pm$2.83}  & 78.50 \tiny{$\pm$3.33} \\ \cline{1-5}
CIS (Llama) & \textbf{92.07} \tiny{$\pm$1.07}  & \textbf{92.26} \tiny{$\pm$0.76}  & \textbf{93.52} \tiny{$\pm$0.73}  & \textbf{92.95} \tiny{$\pm$0.76}  \\ \cline{1-5}
CIS (GPT) (ours) & 88.19 \tiny{$\pm$3.20}  & 90.79 \tiny{$\pm$1.85}  & 89.43 \tiny{$\pm$2.64}  & 86.45 \tiny{$\pm$3.04} 
\\ \cline{1-5}
CIS (classify - GPT) & 88.79 \tiny{$\pm$3.85}  & 89.19 \tiny{$\pm$4.34} &  87.14 \tiny{$\pm$5.38}  & 86.43 \tiny{$\pm$3.30} \\ \hline
\multicolumn{5}{l}{\textbf{Reuters-14}} \\ \hline
CIS (Mistral) & 74.18 \tiny{$\pm$1.29} & 71.86 \tiny{$\pm$4.56}  & 77.27 \tiny{$\pm$1.54}  & 71.73 \tiny{$\pm$8.87} \\ \cline{1-5}
CIS (Llama) & \textbf{83.97} \tiny{$\pm$1.08}  & \textbf{84.61} \tiny{$\pm$1.24}  & 83.27 \tiny{$\pm$1.85}  & \textbf{84.97} \tiny{$\pm$1.15} \\ \cline{1-5}
CIS (GPT) & 82.02 \tiny{$\pm$0.75}  & 82.29 \tiny{$\pm$2.41}  & \textbf{83.94} \tiny{$\pm$1.55}  & 84.53 \tiny{$\pm$1.15} \\ \cline{1-5}
CIS (classify - GPT) & 76.72 \tiny{$\pm$5.08}  & 80.02 \tiny{$\pm$3.53} &  78.00 \tiny{$\pm$4.21}  & 78.61 \tiny{$\pm$3.88} \\ \hline
\end{tabular}
}
\end{center}
\caption{\label{tab:1}Ablation results in \textit{Last Accuracy}. 
}
\end{table}

\noindent
\textbf{Ablation results.} We report ablation results first as they are more interesting. We use the test accuracy after all tasks are learned in Table~\ref{tab:1}, called the \textbf{Last Accuracy}. \textit{Average Incremental Accuracy} and \textit{Forgetting Rate} are also commonly used, but they require accuracy after each task is learned. In the blurry setting, where task boundaries are not clearly defined, these metrics are hard to apply.     

\textbf{\textit{CIS is effective}.}
In the 3/4-blurry, 4/3-blurry, and 5/2-blurry settings,
the accuracy results are comparable to or even exceed (as seen with DBpedia) those obtained when all 7 training samples per class are presented simultaneously (7 Non-blurry). 

{\color{black}\textbf{\textit{Forgetting}.} Comparing the accuracy results of CIS with those of $7$-Non-blurry helps quantify  the amount of \textit{prompt-based forgetting} resulting from incremental summarization. We observe minimal forgetting. For example, for Banking-77 with GPT, forgetting is only about 0.27-0.65\%; for DBpedia-14 with GPT, incremental summary updating ($M/N$-Blurry) even outperforms seeing all data at once ($7$ Non-blurry) by 2-3\%. This may be because LLMs are sensitive to noises, causing in-context learning to be more easily distracted when seeing more samples \cite{shi_why}.

\textbf{\textit{Confidence vs. classification}.} 
In general, using classification in step 2 (\textbf{CIS (classify)}) gives poorer results, though it is slightly better in some cases.

{\color{black}\textbf{\textit{Random order of classes in inference}.} \citet{zheng2024largelanguagemodelsrobust} showed that LLMs are sensitive to the order of options. However, CIS is much less impacted by this because we ask for confidence scores, not for a class.
See details in Appendix~\ref{sec:appendix.random}.
}

}

\vspace{2mm}
\noindent
\textbf{CIS outperforms baselines (Table~\ref{tab:2}).} Here, we use Llama 3.1 as we cannot fine-tune or do Joint prompting on GPT-3.5 due to its small token limit. Compared to the state-of-the-art baseline VAG, CIS dramatically outperforms it with only 7 samples per class. Even when VAG is trained with the full dataset, its performance is considerably below that of CIS. We are aware that this comparison is somewhat unfair, as VAG could not work with recent LLMs.
{\color{black}
However, our CIS is not far from the results of \textit{Joint prompting} and \textit{Joint fine-tuning} using all 7 samples per class, which are often regarded as the upper bounds of CIL. Joint prompting is actually weaker than CIS on the last two datasets. Joint fine-tuning is only better with the full dataset.  
}

}

\section{Conclusion}
This paper made two contributions: (1) proposing a new CL paradigm, called CLOB, which works with black-box LLMs using only verbal prompts. 
(2) proposing a novel CL method CIS based on incremental summarization. Evaluations show that CIS not only has almost no CF but also outperforms baselines by a large margin. 

{
\color{black}
We believe that there are two main reasons for the strong results. First, there is zero model parameter-based catastrophic forgetting, which the baselines have, because they require training or fine-tuning the full or part of the network for each new task. This is a significant advantage of LLMs. By aiming to cover comprehensive knowledge, LLMs can ensure that no essential information is overlooked during summary updates, effectively mitigating the risk of forgetting caused by updating summaries. Second, by using very compact summaries, our classification process can access summaries of all classes or in chunks simultaneously, but the baselines do not have access to the representations of all tasks at the same time as they cannot use summaries. Note that the summary of each class can be seen as its representation. Although replay-based baselines have access to some raw data of earlier classes/tasks, the amount of the raw data is too small for the system to handle parameter-based forgetting.

The experiments conducted in this paper stay within the scope of category-based text classification, as our work focuses on the CIL setting, which requires learning an increasing number of new classes over time. Relation classification is also appropriate for the proposed setting and, thus, a potential future extension of this work.
}

\section{Limitations}\label{sec:limit}

There exists a concern that if each document is too long to fit in the token limit of the LLM, it will be difficult to produce a summary for the document. In such a case, we may need to break the document into multiple chunks and summarize each chunk first. After that, we summarize the chunks of summaries. However, it is unclear whether the approach will be effective. To address this problem, we could also use long-context LLMs.
{\color{black} However, at this point, we cannot draw any conclusion about using very long documents as we do not have such documents. 
New experiments will be needed to test using very long documents to find the most appropriate solution, leading to future work.
}

Another limitation is that the proposed method may be hard to apply to computer vision applications due to the use of summaries. It is unclear how to replace the text summary with some form of image summary. These all form interesting future research directions, which are worthy of exploration because the proposed approach is highly effective and has almost no forgetting, which has plagued the existing fine-tuning or adaptation-based CL.

\section{Ethics Statement}
We believe that our work has no ethical issues or risks as we are using public-domain datasets and our task is simply classification based on the class labels already provided in the datasets. 

\section*{Acknowledgments}
This work was supported in part by four NSF grants (IIS-2229876, IIS-1910424, IIS-1838770, and CNS-2225427) and a research contract from KDDI Research.

\bibliography{custom}

\clearpage
\newpage
\appendix

\section{Prompts Used in CIS}
\label{sec:appendix.prompts}



\subsection{Reflector} 
\framebox[\columnwidth]{\parbox{\columnwidth-3mm}{
I will present you with several examples from the same class. Based on these examples, please provide a summary of the class in no more than \textit{3} sentences. Note that your summary should not include any of the examples.

\textit{Examples:
[list of examples]
}}
}

\subsection{Updator}

\framebox[\columnwidth]{\parbox{\columnwidth-3mm}{
Below is the original summary of a class. I will now provide \textit{m} additional examples within this class. Based on these, please update the summary accordingly. Ensure that the updated summary does not exceed \textit{3} sentences.

\textit{Original summary:
[summary text]}

\textit{Additional examples:
[list of examples]}
}}

\subsection{Solver - Classification}

\framebox[\columnwidth]{\parbox{\columnwidth-3mm}{
Please classify the provided test sample into one of the listed classes based on the summaries provided for these classes. The summaries are formatted as '<class name>: [summary].' Your response should include only one class name as the answer. This name must exactly match one of the classes given. Do not include the original test sample in your response.

\textit{Test sample: [sample]}

\textit{Listed summaries:
<class name>: [summary]}

}}

\subsection{Solver - Confidence}
\framebox[\columnwidth]{\parbox{\columnwidth-3mm}{
Please tell me your confidence scores for the test sample belonging to each class I provided. I will present you a list of summaries for these classes as your reference. The summaries are formatted as '<class name>: [summary].' Your response should include only the class names and the corresponding confidence scores in decimal as the answer. The name must exactly match one of the classes given. Do not include the original test sample in your response.

\textit{Test sample: [sample]}

\textit{Listed summaries:
<class name>: [summary]}

}}





\section{Summary Examples}\label{sec:appendix.examples}


\framebox[\columnwidth]{\parbox{\columnwidth-3mm}{
\textbf{Routing:} "The category is requesting routing numbers for various banks and accounts. These numbers are used to identify the specific bank and account when making transactions. Customers can find their routing numbers through their bank's website or by contacting their bank directly."\\\\
\textbf{Transactions:} "The category is about requesting information on past transactions, including the ability to list recent transactions, check on a specific transaction, and view transaction history within a certain time frame."\\\\
\textbf{Min_payment:} "The category is about minimum payments for various bills, such as truck payments, M\&T bills, power bills, credit card bills, and phone bills. Customers are seeking information on the minimum amount they need to pay for each bill."
}}



\section{Random Order Issue}\label{sec:appendix.random}

As discussed, the order of classes provided to the LLMs affects the results. The top-$k$ confidence scores are no exceptions.

To investigate this issue, we conducted experiments using DBpedia-14 dataset with our CIS system. The experiments using Llama-3.1-8B show the following results:\\

\noindent
3/4-blurry: 90.32 $\pm$0.76\\
4/3-blurry: 90.18 $\pm$3.89\\
5/2-blurry: 90.18 $\pm$1.57\\
Non-blurry: 90.57 $\pm$1.41\\

The results show a slight decline compared to the original results with Llama, with higher standard deviations.
We also ran similar experiments with 3/4-blurry and 4/3-blurry using GPT-3.5, which are reported below:\\

\noindent
3/4-blurry: 92.41 $\pm$0.56\\
4/3-blurry: 93.56 $\pm$0.81\\

In this case, the standard deviations are low, indicating that the class order has a minimal effect on the final classification results.
We also notice that these new results are better than those reported in the paper, but this is possibly due to the update of OpenAI's system.

We investigated the outputs and found that though the top-3 output classes may vary in confidence scores or order, they still appear as the top-3 predictions. This allows our system CIS to pick the correct top-3 class. In some cases, the correct class is no longer the top-1 class, but it is still in the top-3 list.

Below, we show three test samples with their top-3 output classes ranked by confidence. 

\framebox[\columnwidth]{\parbox{\columnwidth-3mm}{
\textbf{Test sample 1:}\\\\
St Nicholas' Church Tuxford is a Grade I listed parish church in the Church of England in Tuxford.\\\\
\textbf{Original top-3 output classes from step 1 in the experiments for the paper:}\\
Top-1: building; 2: village; 3: natural_place\\\\
\textbf{Top-3 output classes from the new experiments with 3 random orderings of classes:}\\
1. Top-1: building; 2: village; 3: natural_place\\
2. Top-1: building; 2: village; 3: natural_place\\
3. Top-1: building; 2: village; 3: natural_place
}}

\framebox[\columnwidth]{\parbox{\columnwidth-3mm}{
\textbf{Test sample 2:}\\\\
Chris Phillips (born March 9 1978) is a Canadian professional ice hockey player for the Ottawa Senators of the National Hockey League (NHL). He has been a member of the Ottawa Senators for his entire career which began with the 1997–98 season. He also serves as their alternate captain and is regarded as a stay-at-home defenceman. The Senators drafted him first overall in the 1996 NHL Entry Draft. He was raised in Fort McMurray Alberta.\\\\
\textbf{Original top-3 output classes from step 1 in the experiments for the paper:}\\
Top-1: athlete; 2: office_holder; 3: company\\\\
\textbf{Top-3 output classes from the new experiments with 3 random orderings of classes:}\\
1. Top-1: athlete; 2: office_holder; 3: company\\
2. Top-1: athlete; 2: company; 3: office_holder\\
3. Top-1: athlete; 2: office_holder; 3: company
}}

\framebox[\columnwidth]{\parbox{\columnwidth-3mm}{
\textbf{Test sample 3:}\\\\
The Himalayan agama (Paralaudakia himalayana) is an agamid lizard found in Central Asia and South Asia.\\\\
\textbf{Original top-3 output classes from step 1 in the experiments for the paper:}\\
Top-1: animal; 2: natural_place; 3: plant\\\\
\textbf{Top-3 output classes from the new experiments with 3 random orderings of classes:}\\
1. Top-1: plant; 2: animal; 3: natural_place\\
2. Top-1: animal; 2: natural_place; 3: village\\
3. Top-1: animal; 2: natural_place; 3: plant
}}

\end{document}